\documentclass[lettersize,journal]{IEEEtran}
\usepackage{amsmath,amsfonts}
\usepackage{algorithmic}
\usepackage{algorithm}
\usepackage{array}
\usepackage[caption=false,font=normalsize,labelfont=sf,textfont=sf]{subfig}
\usepackage{textcomp}
\usepackage{stfloats}
\usepackage{url}
\usepackage{verbatim}
\usepackage{graphicx}
\usepackage{cite}
\usepackage{booktabs}
\usepackage{multirow}
\usepackage{ulem}

\begin{document}
\title{Generalizing monocular colonoscopy image depth estimation by uncertainty-based global and local fusion network}
\author{{Sijia Du, Chengfeng Zhou, Suncheng Xiang, \IEEEmembership{Member, IEEE}, Jianwei Xu, Dahong Qian, \IEEEmembership{Senior Member, IEEE}}
\thanks{This work is supported by the Joint Laboratory of Intelligent Digestive Endoscopy between Shanghai Jiaotong University and Shandong Weigao Hongrui Medical Technology Co., Ltd. 
Corresponding authors: Dahong Qian (dahong.qian@sjtu.edu.cn)
S Du, C Zhou, S Xiang, J Xu, and D Qian are with School of Biomedical Engineering, Shanghai Jiao Tong University, Shanghai, China. (E-mail: Scarlett\_Du@sjtu.edu.cn, chengfengzhou@sjtu.edu.cn, xiangsuncheng17@sjtu.edu.cn, jianwei\_xu@sjtu.edu.cn)
}}

\maketitle

\begin{abstract}
\textbf{Objective:} Depth estimation is crucial for endoscopic navigation and manipulation, but obtaining ground-truth depth maps in real clinical scenarios, such as the colon, is challenging. This study aims to develop a robust framework that generalizes well to real colonoscopy images, overcoming challenges like non-Lambertian surface reflection and diverse data distributions. 
\textbf{Methods:} We propose a framework combining a convolutional neural network (CNN) for capturing local features and a Transformer for capturing global information. An uncertainty-based fusion block was designed to enhance generalization by identifying complementary contributions from the CNN and Transformer branches. The network can be trained with simulated datasets and generalize directly to unseen clinical data without any fine-tuning. 
\textbf{Results: }Our method is validated on multiple datasets and demonstrates an excellent generalization ability across various datasets and anatomical structures. Furthermore, qualitative analysis in real clinical scenarios confirmed the robustness of the proposed method. 
\textbf{Conclusion: }The integration of local and global features through the CNN-Transformer architecture, along with the uncertainty-based fusion block, improves depth estimation performance and generalization in both simulated and real-world endoscopic environments. 
\textbf{Significance:} This study offers a novel approach to estimate depth maps for endoscopy images despite the complex conditions in clinic, serving as a foundation for endoscopic automatic navigation and other clinical tasks, such as polyp detection and segmentation.

\end{abstract}


\section{Introduction}
\label{sec:introduction}
\begin{figure}[htbp]
\setlength{\abovecaptionskip}{0.cm}
\centerline{\includegraphics[width=\columnwidth]{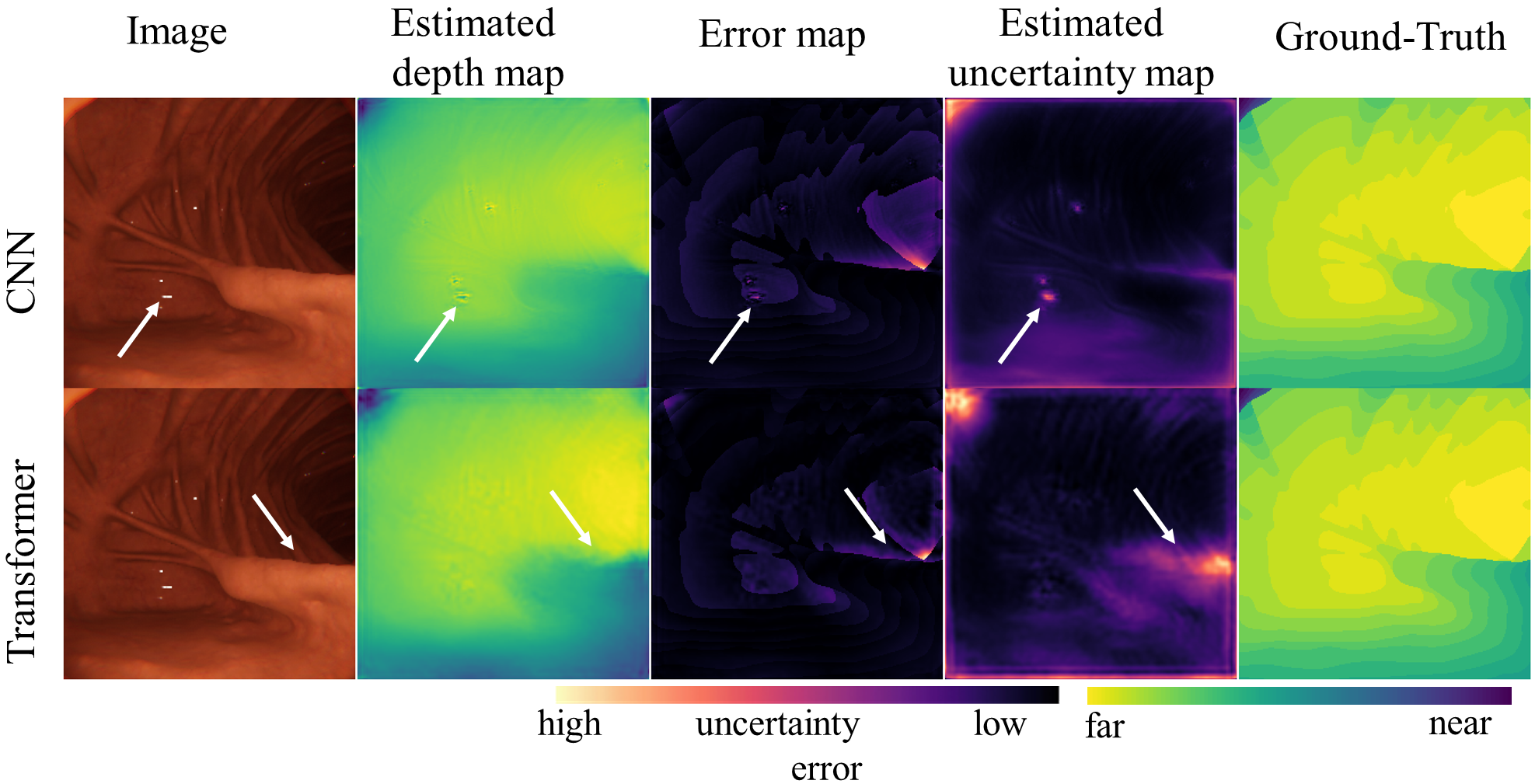}}
\caption{Visualization of the estimated depth map, uncertainty, and error map of CNN and Transformer. Brighter regions in the heatmap indicate larger values. The arrows indicate regions where CNN and Transformer are not proficient at prediction.
The CNN is not adept at predicting reflective regions while the Transformer is not skilled in the depth prediction of structures such as edges, which is complementary and can be shown on the uncertainty maps, respectively. 
}

\label{fig:intro}
\vspace{-0.7cm}
\end{figure}

\IEEEPARstart{E}{ndoscopy} is a conventional diagnostic modality for gastrointestinal diseases, serving as the gold standard for various conditions including gastrointestinal bleeding, polyps, and colorectal\cite{kuipers2005diagnostic}. It is performed with a flexible fiber-optic instrument to visualize and diagnose internal organs or body cavities. Consequently, the task of depth estimation for collected endoscopy images is a fundamental task in scene understanding and also paves the way for endoscopy automatic navigation and diagnosis\cite{fu2021future}. This task can obtain three-dimensional information from two-dimensional images to help physicians better understand the relationship between anatomy, lesions, and organs, thereby obtaining the accurate position of the endoscope. Furthermore, estimating the depth of the lesion is critical for determining the size of the lesion and selecting the appropriate treatment\cite{mahmood2019polyp}.

Since the depth sensors cannot be attached to the endoscope due to space limitations in the colon, depth estimation from monocular endoscopic images is preferred in practice. Compared with previous traditional methods such as shape from shading\cite{RN465} and structure from motion\cite{RN466}, deep learning-based methods\cite{ma2019real,RN497,shao2022self, RN375, RN488, GLP} can estimate depth maps with higher accuracy and robustness. Nevertheless, the lack of ground-truth obtained by depth sensors and the complex environment in the endoscopy video, such as non-Lambertian surface and low lighting, still bring many challenges to these deep learning-based methods. To these ends, self-supervised methods\cite{ma2019real,RN497,shao2022self} were proposed to train the network without ground-truth depth map. However, these methods may encounter challenges in weak illumination and complex colon environments including reflections and air bubbles. Other solutions revolved around image preprocessing and domain adaptation to transfer real images into a simulated data style at either image-level or feature-level. However, these methods cannot be directly applied to depth estimation in clinical data, which may increase the complexity of depth estimation and introduce additional errors.

Furthermore, existing works ignore the complementation between local and global information due to limitations in network structure. For example, CNN-based methods are better at predicting depth maps of the detailed structures because of the local receptive field. However, they are also susceptible to local noise such as reflection and occlusion\cite{oda2022depth}. The Visual Transformer-based methods\cite{RN479,han2022transdssl} have a global receptive field built upon the attention mechanism. Therefore, the Transformer will ignore local noise. However, the detailed features will also be ignored, resulting in unsatisfactory depth estimation results. 

In general, these methods cannot achieve accurate depth maps for images with complex scenes including high-noise regions and low lighting. Consequently, they are not good at simulation-to-real generalization, which needs to process such regions in raw endoscopy images simultaneously. Recently, CNN and Transformer hybrid models have become increasingly popular in visual tasks and demonstrate promising generalization ability and accuracy\cite{bae2023deep,vindas2022hybrid,yuan2023hcformer,nie2022deep}. It inspired us to use the hybrid model to improve the simulation-to-real generalization in colonoscopy depth estimation and achieve more accurate results. However, these hybrid models either lack interpretability\cite{vindas2022hybrid,bae2023deep} or are not suitable for colonoscopy depth estimation tasks in real scenes due to the unexpected noise and distorted features\cite{yuan2023hcformer,nie2022deep}. It is still an unexplored problem to choose a specially designed fusion module to handle the complex scenes in real scenes and improve the simulation-to-real generalization. 

Fortunately, the uncertainty of the predicted depth map serves as a powerful tool to accurately describe the quality of prediction in the abovementioned dynamically changing environment. Uncertainty estimation is a widely-used technique for capturing error distribution and is extensively applied in real-world scenarios, particularly in high-risk domains \cite{karimi2019accurate}. As shown in Fig. \ref{fig:intro}, the CNN and Transformer have different errors in the same regions because of their different areas of expertise. Additionally, the regions with high uncertainty is highly consistent with the regions with large depth errors. Therefore, we propose a novel hybrid CNN-Transformer framework, which combines the reliable parts of the two networks' predictions using an uncertainty-based fusion module. This module estimates the uncertainty map for each branch in an unsupervised learning manner and employs it to determine the trustworthy parts of the two predictions. It effectively mitigates the poor depth estimation in endoscopy, where highly dynamic scenes, such as air bubbles, reflection, and insufficient lighting are prevalent.

In conclusion, our contributions can be summarized as:

\begin{enumerate}
    \item We propose a novel monocular depth estimation framework that combines CNN and Transformer to make full use of the complementary features to estimate depth maps with higher accuracy.

    \item We design a novel uncertainty-based fusion module to improve the effective fusion of local and global features. The uncertainty maps are estimated in an unsupervised manner and could allow greater reliability for the endoscopy scenario. 

    \item Our method is validated on multiple datasets and demonstrates an excellent generalization ability across various datasets and anatomical structures. Furthermore, our method can directly perform depth estimation on real colonoscopy images without complex preprocessing and domain adaptation.
\end{enumerate}

\section{Related Work}
\subsection{Monocular Depth Estimation for Endoscopic Images}

Leveraging the assumption of illumination consistency, many studies conduct self-supervised depth estimation, where they jointly estimate the depth map and camera pose by minimizing the photometric loss. For instance, Ma et al.\cite{ma2019real} used a recurrent neural network based on illumination consistency to construct surfaces for colon chunks with a predicted depth map. Ozyoruk et al.\cite{RN497} proposed Endo-SfMLearner with a brightness-aware photometric loss to improve the assumptions of illumination consistency and make it work in colonoscopy images. Shao et al.\cite{shao2022self} introduced the appearance flow to address the brightness inconsistency problem. However, due to the absence of ground-truth, these self-supervised methods often yield low-accuracy sparse depth maps. Furthermore, these methods may encounter challenges such as reflections and the presence of non-informative frames in the clinic, leading to potential failures in depth estimation. 

Conversely, the dense depth ground-truth is readily available for simulated endoscopic images. Thus, many studies have conducted supervised training using simulated data and bridged the gap between simulation and reality with data preprocessing or domain adaptation. Part of these methods\cite{RN461,RN463,RN475} initially generated synthetic images with ground-truth from CT scans, then trained the model with the paired synthetic data. During their test phase, a generative adversarial network (GAN) was used to transfer the real colonoscopic images into synthetic styles, which were then formatted as inputs for the trained model. Other research\cite{RN477,RN476} employs the opposite process, which maps the real images' texture to synthetic images using the CycleGAN for model training, and directly inputs the real image to the trained model. In general, the former methods may result in low-accuracy due to style transfer errors while the latter method is difficult to extend to general medical images due to the complex texture of human tissue. Nevertheless, the above methods require complex process before depth estimation, which increase the difficulty of applying depth estimation to clinical images. In this paper, we are the first to present a novel approach that enables trustworthy depth map estimation for raw endoscopy images using an model trained solely with simulated images.

\begin{figure*}[htbp]
\setlength{\abovecaptionskip}{0.cm}
\centerline{\includegraphics[width=2\columnwidth]{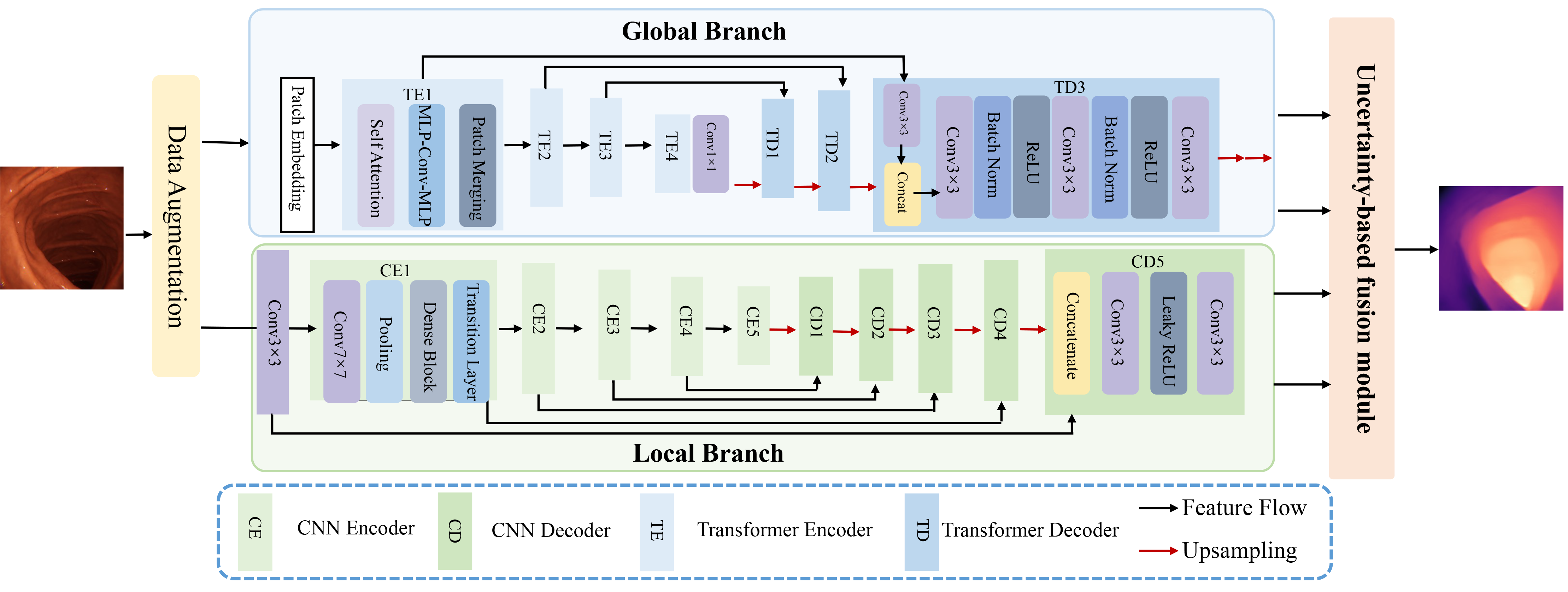}}
\caption{The structure of the proposed colonoscopy depth estimation network. It consists of a local branch, a global branch, and an uncertainty-based fusion module. Local information is extracted using CNN, while global features are emphasized by Transformer. Finally, the uncertainty-based fusion module refines the predictions from two branches using uncertainty maps.}
\label{fig:network}
\vspace{-0.7cm}
\end{figure*}
\vspace{-0.2cm}

\subsection{Fusion of CNN and Transformer}
Understanding both global and local information is crucial for depth estimation, especially for the complex and changeable environment of colonoscopy. The hybrid model of CNN and Transformer is a promising tool to handle the complementary features simultaneously. Intuitively, Cao et al.\cite{cao2022cnn} used a feature-focus sub-network realized by channel-wise concatenation and addition to fuse the local and global features of CNN and Transformer. However, this method cannot highlight the characteristics of different features, thereby limiting the ability of the hybrid model. Vindas et al.\cite{vindas2022hybrid} added the attention mechanism into the feature fusion operation and proposed a late fusion module, where the attention map is used to weigh the signal representation of CNN and Transformer. Bae et al.\cite{bae2023deep} extended this idea by adapting the Attention Connection Module, which generates the position and channel attention maps to capture local details and shape biases within the CNN and Transformer.  
In addition, there are some works which fuse CNN and Transformer into a single network. Yuan et al.\cite{yuan2023hcformer} replaced the multi-layer perceptron (MLP) with CNN in the Transformer, while Nie et al.\cite{nie2022deep} employed CNN to extract local feature maps from the images, which were then used as inputs to the Transformer for global feature modeling. Compared with methods with explicit fusion modules, these methods lack interpretability to some extent. By contrast, our method adopts an uncertainty-based module with fewer computing resources but stronger interpretability. This mechanism is more suitable for colonoscopy images with various noises because the uncertainty map can intuitively evaluate the error distribution of the depth map. In addition, the difference in the distribution of two uncertainty maps of the two branches can make full use of the respective advantages of CNN and Transformer in an interpretable way.

\subsection{Uncertainty Estimation for Monocular Depth Estimation}

Uncertainty estimation is a useful tool to reflect the error distribution and is widely used in the medical field. It can improve the reliability of medical diagnosis\cite{karimi2019accurate} and the generalization ability of the model between different datasets\cite{gong2019icebreaker}. Depth estimation is the basis of endoscopic navigation techniques. Therefore, it is critical to identify regions with higher errors to avoid catastrophic results.
Introducing uncertainty estimation in the colonoscopy depth estimation task can effectively filter out regions with large errors, thereby ensuring the safety and robustness of subsequent 3D reconstruction and navigation tasks. 
Poggi et al.\cite{RN272} first introduced uncertainty to the monocular depth estimation task and compared the impact of different uncertainties on the accuracy of depth estimation. Rodriguez et al.\cite{RN277} used Bayesian deep networks for single-view depth estimation in colonoscopies. 
Most of these works\cite{RN224, RN279} stop at estimating the uncertainty of the depth map, but few of them further use uncertainty to improve the prediction accuracy and generalization ability of the network. In this work, we explore for the first time uncertainty estimation for CNN and Transformer feature fusion to fully utilize the uncertainty in colonoscopy depth estimation.

\section{Methods}
In this section, we present depth estimation methods that utilize an uncertainty-based fusion module to handle both CNN and Transformer information, enabling the prediction of the depth map $\hat{D} \in \mathbb{R}^{H \times W \times 1}$ for the monocular colonoscopy image $I \in \mathbb{R}^{H \times W \times 3}$.
 The overall network architecture, which includes a CNN branch, a Transformer branch, and an uncertainty-based fusion module, is illustrated in Fig. \ref{fig:network}. 
 After resizing and data augmentation, the RGB image $I$ is fed into both the global and local branches to obtain two depth maps. These predictions are then passed into the uncertainty-based fusion module, which leverages uncertainty maps to achieve more accurate results.
 In the following sections, we describe the detailed network structure of the two branches and explain how the uncertainty and depth maps are predicted simultaneously, with the uncertainty map obtained in an unsupervised manner. Additionally, we introduce the uncertainty-based fusion module's detailed structure and loss functions to support the network training.
Furthermore, we clarify the data augmentation methods specifically designed for colonoscopy depth estimation, aiming to improve the network's generalization ability.

\begin{figure}[htbp]
\setlength{\abovecaptionskip}{0.cm}
\centerline{\includegraphics[width=\columnwidth]{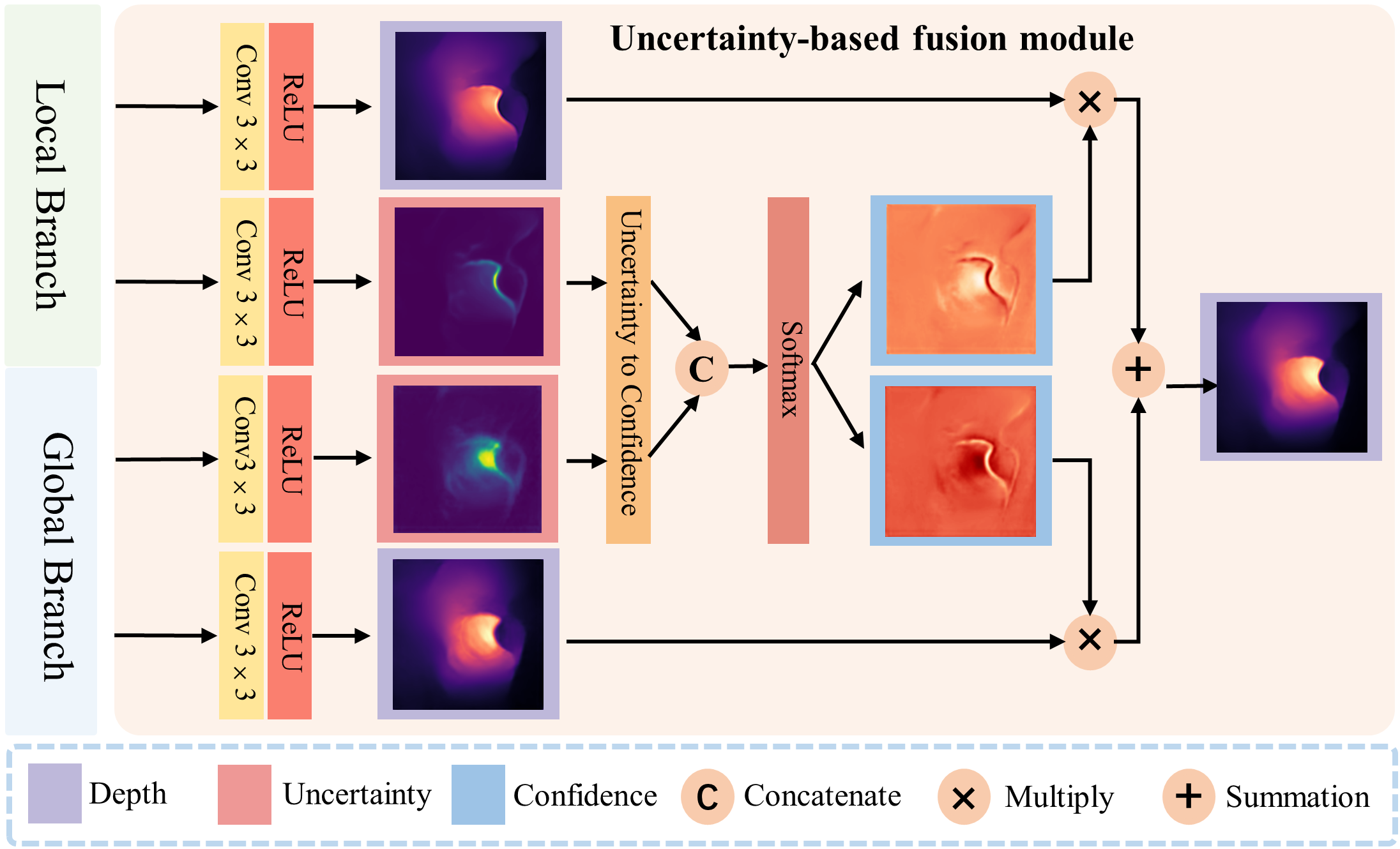}}
\caption{The topology of the proposed uncertainty-based fusion module. Layers in global and local branches pass through convolutions for uncertainty estimation. The obtained uncertainty maps are transformed into confidence maps and passed through a Softmax layer to weigh the respective depth maps. The weighted results of the branches are then combined for the final output.}
\label{fig: uncertainty-based fusion module}
\vspace{-0.6cm}
\end{figure}

\subsection{Local Branch and Global Branch}
To fully extract local and global information from the image respectively, we employ a simple yet effective two-branch approach. In the local branch, we utilize a CNN with a local receptive field, achieved through convolutional operations. Meanwhile, the global branch utilizes a Transformer with a self-attention mechanism, enabling it to capture global receptive field information.

The local branch is a concise CNN inspired by Alhashim \cite{RN375}, using the DenseNet-169 \cite{RN481} pretrained on ImageNet \cite{RN482} as its encoder block to gradually scale
 the feature maps to $8\times8$. Compared to ResNet \cite{resnet}, Dense Convolutional Network serves as a better local feature extractor with fewer parameters. The decoder block has a simple structure, consisting of two convolution layers with $3 \times 3$ kernel sizes and bilinear upsampling layers to increase the spatial resolution of the estimated depth map. Additionally, skip-connections are applied to integrate feature maps from both the encoder and decoder blocks. In general, the local branch adopts a five-layer encoder-decoder structure to fully utilize local features. 

Similarly, a simple Transformer-based global branch is derived for powerful global feature extraction inspired by \cite{GLP}. It comprises of a patch embedding block, Transformer encoder blocks proposed by Xie et al.  \cite{RN483} and simple decoders.  
The patch embedding block converts input RGB images into sets of $3 \times 3$ and $7 \times 7$ vectors. These embedded patches are then fed into a simple but efficient Transformer encoder block, providing global insight into the RGB image features. The encoder block consists of a self-attention module and lightweight MLP layers. Dropout is applied in the MLP with a probability of $0.5$ to enhance the model's generalization ability. 
The features from the bottom encoder are subsequently fed through 1 $\times$ 1 convolution layers and decoder blocks to be integrated with features from other encoder blocks, where the decoder block involves $3 \times 3$ convolution layers with batch normalization and ReLU activation.

Finally, each last layer of the two network branches needs to pass through the convolutional layer and the activation layer to obtain their respective depth estimation maps.

\subsection{Uncertainty-based Fusion Module}
To solve the issues affecting endoscopy depth estimation including air bubbles, reflection, and edges blur simultaneously, we propose an uncertainty-based fusion module to fuse the depth maps of the two branches using the complementation between CNN and Transformer. These errors can be reflected by the uncertainty maps of both branches. The confidence map derived from these uncertainty maps allows the network to account for various error factors during prediction, thus addressing multiple sources of depth estimation errors and enhancing overall robustness.

Firstly, to simultaneously estimate the depth and uncertainty maps in a single branch, we extract the maximum posterior (MAP) loss function from a probabilistic depth estimation problem. Additionally, a concise structure is introduced to automatically determine the trustworthy parts of the predictions from the two branches based on their respective uncertainty maps. Overall, the uncertainty-based fusion module does not need additional training steps for uncertainty estimation\cite{RN272} and has strong interpretability.

\subsubsection{Uncertainty Estimation for Depth Predictions}
Unlike previous works that directly learn the  point estimation of depth maps $\hat{D}$ from input images $I$ as $\hat{D} = F(I)$, we treat the depth estimation task as an interval estimation problem that considers the distribution of prediction errors, i.e., uncertainty, while predicting the depth map. To enable end-to-end training and application of the network, 
it is essential for the network to simultaneously predict depth and uncertainty. Thus, we further solve the depth and uncertainty simultaneously by MAP methods \cite{RN272}, taking into account the input images, the estimated depth maps, and the uncertainty maps of the predictions. This probabilistic formulation can transform into a loss function during training, optimizing both depth and uncertainty predictions. This approach proves to be more efficient compared to other uncertainty estimation methods that require additional training steps \cite{MCDrop, ensemble}, making it well-suited for depth estimation tasks in endoscopic navigation and clinical examinations.

The joint posterior probability $P$ of the predicted depth map $\hat{D}$, the uncertainty map $\Sigma$, and the input image $I$ can be formulated as $P\left(\hat{D}, \Sigma|I\right)$. When the posterior probability reaches its maximum, the optimal depth estimate can be obtained along with its corresponding uncertainty map.
\begin{equation}
    \hat{D}=\operatorname{argmax} P\left(\hat{D}, \Sigma|I\right). \label{2}\\
\end{equation}

Furthermore, by denoting the pixel-wise element of $I$, $\hat{D}$ and  $\Sigma$ as $i_j, \hat{d}_j, \sigma_j$, respectively. We can express the probability $P$ as 
\begin{equation}
\begin{aligned}
     & P\left(\hat{D}, \Sigma|I\right) \\
   &= P\left(\Sigma|I) P(\hat{D}|\Sigma,I\right) \\
    &= \underset{j \in I}{\Pi}  p\left(\sigma_j|i_j\right) p\left(\hat{d}_j|\sigma_j, i_j\right). \\
 \end{aligned}
\end{equation}

By taking the logarithm of the above equation, we can transform the maximum likelihood problem of equation (2) into
\begin{equation}
    \max \underset{j \in I}{\Sigma}log\left( p\left(\sigma_j \mid  i_j\right)\right) + log\left( p\left(\hat{d}_j \mid \sigma_j, i_j\right)\right).
\end{equation}

The first term of the equation is the likelihood of the uncertainty map, which can be considered as Jeffrey’s prior\cite{RN495}. The latter term can then be modeled as a Gaussian distribution according to Jakob et al.\cite{RN275}. Finally, the depth map is estimated by minimizing the following  MAP function, where $d_j$ denotes the pixel-wise element of the ground-truth depth $D$:
\begin{equation}
    \begin{aligned}
         \underset{j \in I}{\sum}\left( 2 \log \sigma_j+\frac{\left(\hat{d}_j-d_j\right)^2}{2 \sigma_j^2}\right).
    \end{aligned}
\end{equation}

By incorporating the above loss function into the training process, we can simultaneously estimate the uncertainty and depth maps. Regions with higher uncertainty can indicate regions where the depth estimation error is significant.

\subsubsection{Global and Local Branch Fusion with Uncertainty Map}
In our uncertainty-based fusion module, we incorporate an independent convolutional operation to compute the uncertainty for each branch. As shown in Fig. \ref{fig: uncertainty-based fusion module}, the layers of each branch is passed through a $3\times3$ convolutional layer with ReLU non-linear activation for uncertainty estimation.

To further refine the uncertainty estimation, we convert the estimated uncertainty map into a confidence map inspired by \cite{uuu}, assigning lower weights to regions with higher uncertainty. The confidence map is denoted as $C$ and  derived by
\begin{equation}
    C = Sigmoid\left(\exp \left( -\Sigma\right)\right).
\end{equation}

After that, the confidence maps of the two branches are inputted into the Softmax layer to obtain a representation of the error probability distribution
\begin{equation}
\begin{aligned}
    &C^{global'} = \frac{e^{C^{global}}}{e^{C^{global}}+e^{C^{local}}},\\
    &C^{local'} = \frac{e^{C^{local}}}{e^{C^{global}}+e^{C^{local}}}.
\end{aligned}
\end{equation}

Here the $C^{global}$ and $C^{local}$ mean the confidence map of the predicted depth map in the global and local branch, respectively.  The $C^{global'}$ and $C^{local'}$ denote the output after passing $C^{global}$ and $C^{local}$ through the Softmax layer. The base $e$ is used to raise these values in the Softmax function. The depth map estimated by the fusion module is expressed as a weighted sum of local and global branches based on the above confidence maps, so as to fully utilize the complementary information between the two branches. Finally, the estimated depth map $\hat{D}$ is noted as
\begin{equation}
    \hat{D} = \hat{D}^{global} \cdot  C^{global'} + \hat{D}^{local} \cdot  C^{local'}.
\end{equation}

\begin{table*}[htbp]
  \centering
  \caption{Ablation study on the EndoSLAM dataset.The
best results are in bolded, and the second-best ones are underlined. The $\uparrow$ means the higher is better, while the $\downarrow$ means the lower is better.}
\scalebox{0.7}{
    \begin{tabular}{cccccccccccc}
    \toprule
          &       &       & $\delta_1\uparrow$    & $\delta_2\uparrow$    & $\delta_3\uparrow$    & abs\_rel$\downarrow$ & sq\_rel$\downarrow$ & RMSE$\downarrow$  & RMSE\_log$\downarrow$ & log10$\downarrow$ & silog$\downarrow$ \\
    \midrule
    \multirow{4}[1]{*}{single branch} & \multirow{2}[1]{*}{CNN} & w/o uncert & 0.857$\pm$0.006	&0.976$\pm$0.002	&0.993$\pm$0.0	&0.129$\pm$0.002	&0.146$\pm$0.005	&0.891$\pm$0.007	&0.2$\pm$0.009	&0.055$\pm$0.001	&19.102$\pm$0.923

  \\
    & &   w uncert  & 0.851$\pm$0.015	&0.973$\pm$0.005	&0.991$\pm$0.002	&0.133$\pm$0.008	&0.161$\pm$0.019	&0.938$\pm$0.049	&0.186$\pm$0.014	&0.056$\pm$0.003	&17.719$\pm$1.432

  \\
    \cmidrule{2-12}
    & \multirow{2}[1]{*}{Transfomer} & w/o uncert & 0.854$\pm$0.007	&0.96$\pm$0.003	&0.978$\pm$0.001	&0.14$\pm$0.004	&0.168$\pm$0.008	&0.875$\pm$0.01	&0.524$\pm$0.049	&0.079$\pm$0.003	&51.3$\pm$5.043

  \\
    & & w uncert & 0.85$\pm$0.008	&0.967$\pm$0.006	&0.987$\pm$0.004	&0.139$\pm$0.005	&0.181$\pm$0.02	&0.946$\pm$0.045	&0.264$\pm$0.09	&0.062$\pm$0.009	&25.459$\pm$8.795

  \\
     \midrule
    \multirow{8}[3]{*}{two branches} 
   & \multirow{2}[1]{*}{branch ablations} & cnn$+$cnn & 0.895$\pm$0.001	&0.985$\pm$0.002	&0.997$\pm$0.0	&\uline{0.109$\pm$0.0}	&0.101$\pm$0.002	&0.776$\pm$0.008	&0.152$\pm$0.001	&0.047$\pm$0.0	&14.001$\pm$0.171

  \\
   & & Transformer$+$Transformer &\uline{0.897$\pm$0.005}	&0.983$\pm$0.001	&0.995$\pm$0.001	&0.11$\pm$0.003	&0.111$\pm$0.01	&0.758$\pm$0.019	&0.158$\pm$0.004	&\uline{0.046$\pm$0.001}	&14.731$\pm$0.412
  \\
  & \multirow{2}[1]{*}{fusion module ablations} & concat &0.894$\pm$0.002	&\textbf{0.987$\pm$0.001}	&\textbf{0.997$\pm$0.0}	&0.11$\pm$0.0	&\textbf{0.097$\pm$0.001}	&0.752$\pm$0.004	&0.15$\pm$0.004	&0.047$\pm$0.0	&13.663$\pm$0.365
  \\
   & & merge equally &0.895$\pm$0.007	&0.985$\pm$0.003	&0.997$\pm$0.001	&0.11$\pm$0.004	&0.098$\pm$0.007	&\uline{0.75$\pm$0.034}	&\uline{0.149$\pm$0.002}	&0.047$\pm$0.001	&\uline{13.634$\pm$0.333}
  \\
  
\cmidrule{2-12}          & \multirow{4}[2]{*}{uncertainty-based fusion}& w/o MAP Loss &0.889$\pm$0.006	&0.98$\pm$0.005	&0.993$\pm$0.006	&0.115$\pm$0.005	&0.114$\pm$0.012	&0.771$\pm$0.012	&0.275$\pm$0.187	&0.055$\pm$0.013	&26.481$\pm$18.692

  \\
& & local branch &0.893$\pm$0.003	&0.983$\pm$0.003	&0.996$\pm$0.001	&0.109$\pm$0.001	&0.103$\pm$0.005	&0.784$\pm$0.019	&0.164$\pm$0.003	&0.047$\pm$0.001	&15.231$\pm$0.243

\\
          &       & global branch &0.893$\pm$0.003	&0.982$\pm$0.002	&0.995$\pm$0.0	&0.112$\pm$0.002	&0.118$\pm$0.002	&0.772$\pm$0.012	&0.163$\pm$0.002	&0.047$\pm$0.001	&15.246$\pm$0.198

  \\
          &       & \textbf{final output} &\textbf{0.903$\pm$0.002}	&\uline{0.985$\pm$0.002}	&\uline{0.996$\pm$0.0}	&\textbf{0.106$\pm$0.001}	&\uline{0.098$\pm$0.001}	&\textbf{0.742$\pm$0.01}	&\textbf{0.146$\pm$0.002}	&\textbf{0.045$\pm$0.0}	&\textbf{13.446$\pm$0.229}

          \\
    \bottomrule
    \end{tabular}}
  \label{tab:ablation}%
\end{table*}%

\begin{figure*}[htbp]
\centerline{\includegraphics[width=1.8\columnwidth]{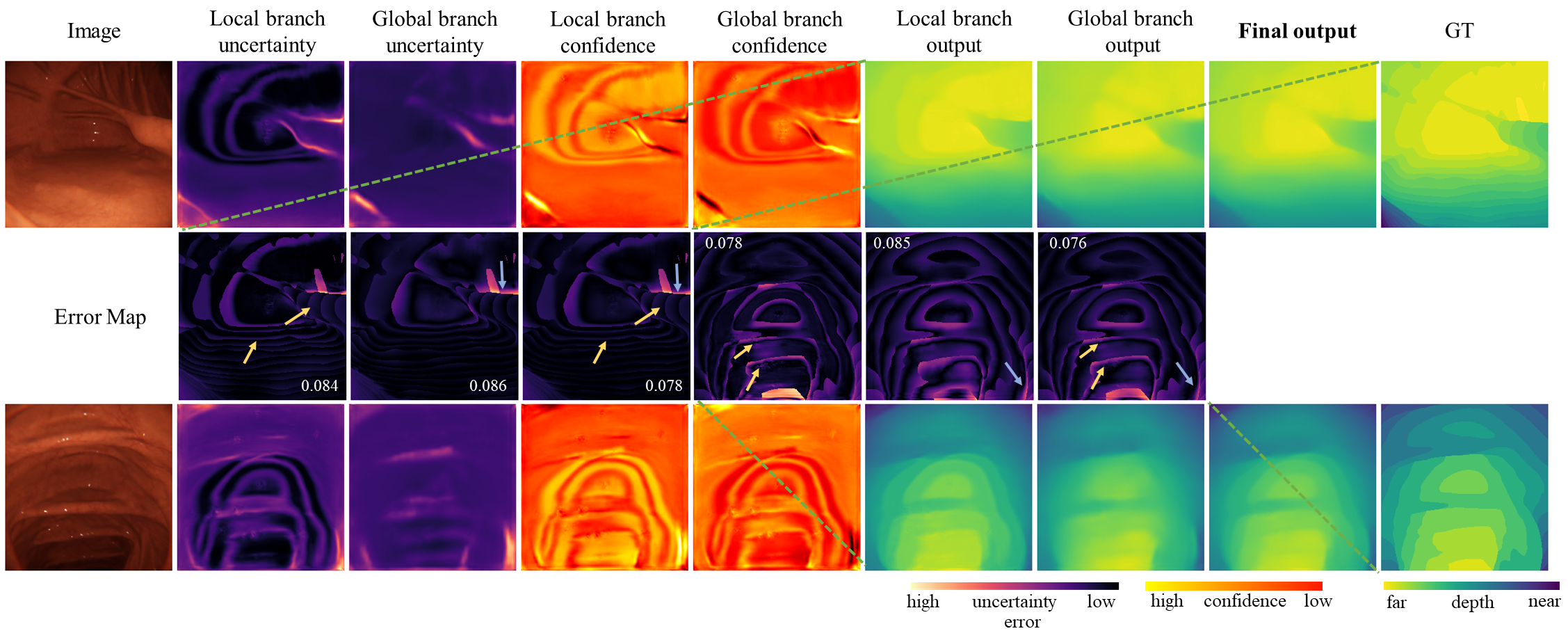}}
\setlength{\abovecaptionskip}{0.cm}
\caption{The intermediate output of the network. The brighter the color, the larger the uncertainty, confidence, and depth value. The arrows highlight the regions where the predicted depth maps are refined according to the uncertainty maps, where the yellow ones represent corrections based on CNN and the blue ones represent Transformer-based corrections. }
\label{fig:ablation explain}
\vspace{-0.7cm}
\end{figure*}
\subsection{Loss Function}

The loss function consists of two parts to ensure that the preliminary predictions of the two branches $\hat{D}^{global}$ and $\hat{D}^{local}$ can be fused to get a better depth map $\hat{D}$. Firstly, the MAP Losses are used to constrain the depth estimation and uncertainty estimation of the two branches, and other loss functions are used to constrain the fused depth map.
\subsubsection*{MAP Loss}
Firstly, the MAP Loss constructed according to equation (4) is used to obtain the depth and uncertainty map of a single branch simultaneously, where the uncertainty map is obtained in an unsupervised manner. The subscript $b$ means the global branch or local branch.
\begin{equation}
L_{MAP}^b = \frac{1}{n} \sum\limits_{j \in I}^{n} \frac{\lvert \hat{d_j}-d_j\rvert}{\sigma_j} + \lambda_b \cdot  log \left( \sigma_j \right).  
\end{equation}

The first term can make the difference between the prediction and ground-truth as small as possible while obtaining the largest uncertainty. The second term can ensure that the uncertainty map will not be infinite which may lead to the wrong prediction. Both the $\lambda_{global}$ and $\lambda_{local}$ are set to 0.1 based on empirical observations in our experiments. 

Since the MAP Loss can only constrain the depth estimation of a single branch, additional loss functions need to be introduced to ensure the accuracy of the fused depth map. Therefore, we have introduced additional loss functions, namely the \textbf{Depth Loss}, and the \textbf{Edge Loss} to penalize the distortion of the fused depth map. 

\subsubsection*{Depth Loss}
The depth loss is the most commonly used loss in the depth estimation task to penalize the value difference between the estimated depth map $\hat{D}$  and the ground-truth $D$. It is expressed as the point-wise L1 loss between $\hat{D}$ and $D$.
\begin{equation}
    L_{depth} = \frac{1}{n} \sum\limits_{j \in I}^{n} \lvert \hat{d_j} - d_j\rvert.
\end{equation}

\subsubsection*{Edge Loss}
The gradient constraint is necessary to ensure a clear boundary of the predicted depth map, where the $g_x$ and $g_y$ mean the element-wise difference of the depth map in the x and y directions respectively. 
\begin{equation}
L_{edge}=\frac{1}{n}  \sum\limits_{j \in D}^{n}\left|\boldsymbol{g}_{\mathbf{x}}\left(d_j, \hat{d}_j\right)\right|+\left|\boldsymbol{g}_{\mathbf{y}}\left(d_j, \hat{d}_j\right)\right|.
\end{equation}

Lastly, the total loss is derived from the weighted sum of all the loss functions mentioned above:

\begin{equation}
\begin{aligned}
       &L = \lambda_{global} \cdot L_{MAP}^{global} + \lambda_{local} \cdot L_{MAP}^{local} \\
    &+\lambda_{depth} \cdot L_{depth}  + \lambda_{edge} \cdot L_{edge}.
\end{aligned}
\end{equation}

\subsection{Data Augmentation}
Due to the tubular structure of the colon, the data augmentation methods suitable for the colonoscopy depth estimation task differ from those commonly used in autonomous driving tasks, where geometric transformations such as horizontal flips and cropping are prevalent.
For example, we found that vertical flip and image rotation, which are forbidden in indoor scenes or street scenes, are all effective for colonoscopy images. Furthermore, the texture of the colon varies in different color channels, so random color channel permutations can further improve the  generalization ability. 
Therefore, we consider all these data augmentation methods at a probability of $0.5$ to boost the performance of depth estimation.

\section{Experiments and Results}

\subsection{Implement Details}
The algorithm was primarily implemented using the PyTorch (version 1.12.1) framework in Python 3.7.6. The training process was accelerated using an NVIDIA GeForce RTX 3080Ti GPU. Before training, all images were crop off the black borders, with size of $256 \times 256$. Initially, the two branches were pre-trained independently. Subsequently, the entire network was fine-tuned for $25$ epochs, using a batch size of $10$. 
We employed the Adam optimizer for updating the network parameters.
The learning rate is adjusted using the ExponentialLR scheduler with initial learning rate of $1e-4$ and $\gamma$ of 0.9. 
The hyper-parameters $\lambda_{global}$, $\lambda_{local}$, $\lambda_{depth}$, and $\lambda_{edge}$ were empirically set to $0.1$, $0.1$, $1.0$,  and $1.0$.

The evaluation of the algorithm was conducted as follows. Firstly, the model was trained on the colon part of the \textbf{EndoSLAM Dataset}\cite{RN497}, which comprises of simulated images of the colon, small intestine, and stomach with depth ground-truth. The dataset reflects the luminosity variations such as the regions with edges and low lighting and reflections that may occur in clinical settings. 
The synthetic images were generated using the VRCaps simulation environment.  For images with black borders, we crop off the black borders on all sides, resulting in the image size becoming 256$\times$256.
Subsequently, the trained model was tested on not only the other anatomical structures in the EndoSLAM Dataset but also two out-of-distributioned simulated datasets with depth ground-truth and colon images rendered using different methods.
The \textbf{Colondepth Dataset}\cite{RN474} was rendered in the Unity game engine and contains $16,016$ RGB images with corresponding ground-truth, all with a resolution of 256$\times$ 256. The \textbf{Scenario Dataset}\cite{RN486} was generated from CT and rendered with Blender, containing $10,000$ images, each with a resolution of $320\times320$. All the images are resized to 256 $\times$ 256 during training and testing.
We conduct experiences under multi different seeds. The final results are represented as the mean$\pm$standard deviation.

\begin{figure*}[htbp]
\setlength{\abovecaptionskip}{0.cm}
\centerline{\includegraphics[width=1.8\columnwidth]{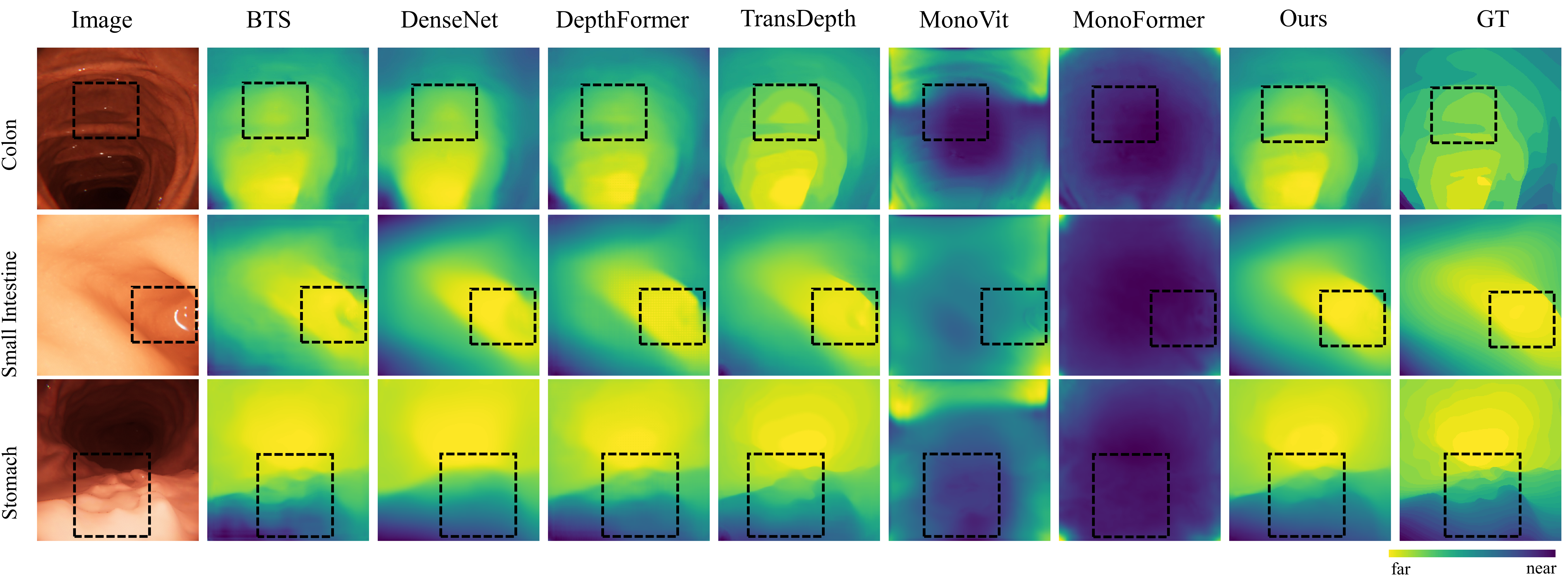}}
\caption{Qualitative comparison of different methods on the EndoSLAM dataset with different anatomical structures. Black boxes highlight regions where our method achieves better predictions. Our method can reconstruct more detailed depth maps without being affected by reflective regions.}
\label{fig:endoslam}
\vspace{-0.3cm}
\end{figure*}

For evaluation metrics, we adopted the generic evaluation metrics proposed by Eigen et al.\cite{RN410}, consisting of nine indicators: $\delta_1$, $\delta_2$, $\delta_3$, abs\_rel, RMSE, RMSE\_log, sq\_rel, silog, and log10. A scaling factor is calculated between the ground-truth and the predictions to scale the predicted depth
maps before evaluation\cite{RN410}.
\begin{equation}
    scale = \frac{\text{median}\left(D\right)}{\text{median}\left(\hat{D}\right)}
\end{equation}
Higher values of $\delta_1$, $\delta_2$, and $\delta_3$ indicate higher accuracy in depth estimation. Conversely, for the remaining indicators, smaller values indicate better results with smaller error.

\subsection{Ablation Study}
We conducted an ablation study on the EndoSLAM dataset to demonstrate the effectiveness of each module. First, we evaluated the performance of the CNN and Transformer baselines, both with and without the uncertainty prediction branch. 
Second, to assess the effectiveness of fusing CNN and Transformer, we experimented by setting both branches to CNN and both to Transformer. 
Additionally, to validate the superiority of our fusion module, we replaced it with other operations: \textit{concat}, which concatenates and convolves the outputs of the two branches, and \textit{merge equally}, which sums and averages the outputs during joint training. 
We also performed an ablation study on the proposed MAP Loss to evaluate its contribution to performance improvements. Finally, we present the intermediate outputs from the two branches and the final output to illustrate the working principles of our method.

As demonstrated in Tab. \ref{tab:ablation}, when the network combines two branches, its performance is significantly improved compared with the single branch version. 
Additionally, the dual-branch model combining CNN and Transformer outperformed the CNN+CNN and Transformer+Transformer dual-branch models, indicating that the fusion of local and global information can more effectively enhance the network's performance.
As for the uncertainty-based fusion module, our method outperforms other fusion methods with a significant improvement of $\delta_1$. It denoted that the uncertainty-based fusion module can better utilize the information extracted from the two branches. Furthermore, results also show that our proposed MAP Loss can improve the accuracy of depth estimation.

To further illustrate the working principle of our proposed method, we present visualizations of the intermediate output of two branches and the final output in Fig. \ref{fig:ablation explain}. It can be seen that the reflection and comparatively flatter regions will affect the prediction of the local branch because accurate prediction in these areas requires more global information as reference. Conversely, the global network branch has a larger receptive field, resulting in less attention to reflection, but struggles with the prediction of edges.
The uncertainty-based fusion of the depth maps generated by the two branches can effectively use the respective advantages of the two branches to improve the accuracy of depth estimation.

\begin{table}[htbp]
\setlength\tabcolsep{2pt} 
  \centering
  \caption{Quantitative evaluation results of different methods using the EndoSLAM dataset. The best results are in bolded, and the second-best ones are underlined.}
    \resizebox{\linewidth}{10mm}{
    \begin{tabular}{ccccccccccc}
    \toprule
    \multicolumn{2}{c}{methods} & $\delta_1\uparrow$    & $\delta_2\uparrow$    & $\delta_3\uparrow$    & abs\_rel$\downarrow$ & sq\_rel$\downarrow$ & RMSE$\downarrow$  & RMSE\_log$\downarrow$ & log10$\downarrow$ & silog$\downarrow$ \\
    \midrule
    \multicolumn{2}{c}{BTS} &0.888$\pm$0.007	&0.984$\pm$0.001	&\uline{0.996$\pm$0.001}	&0.119$\pm$0.007	&0.438$\pm$0.57	&0.827$\pm$0.04	&0.174$\pm$0.026	&0.05$\pm$0.003	&16.324$\pm$2.602

  \\
    \multicolumn{2}{c}{DenseDepth} &0.816$\pm$0.005	&0.953$\pm$0.001	&0.977$\pm$0.002	&0.163$\pm$0.004	&0.219$\pm$0.01	&1.056$\pm$0.024	&0.226$\pm$0.007	&0.065$\pm$0.001	&21.713$\pm$0.718

  \\
    \multicolumn{2}{c}{DepthFormer} &0.883$\pm$0.003	&0.982$\pm$0.0	&0.996$\pm$0.0	&0.118$\pm$0.001	&0.118$\pm$0.001	&0.83$\pm$0.002	&0.156$\pm$0.001	&0.049$\pm$0.0	&14.565$\pm$0.132

  \\
    \multicolumn{2}{c}{TransDepth} &\uline{0.9$\pm$0.005}	&\uline{0.984$\pm$0.003}	&0.995$\pm$0.002	&\textbf{0.105$\pm$0.004}	&\uline{0.101$\pm$0.004}	&\uline{0.754$\pm$0.013}	&\uline{0.146$\pm$0.003}	&\textbf{0.044$\pm$0.001}	&\textbf{13.284$\pm$0.371}

  \\
  
    \multicolumn{2}{c}{MonoVit} &0.284$\pm$0.001	&0.549$\pm$0.005	&0.755$\pm$0.006	&0.591$\pm$0.003	&2.091$\pm$0.022	&3.185$\pm$0.034	&0.596$\pm$0.006	&0.206$\pm$0.002	&58.846$\pm$0.619

  \\
    \multicolumn{2}{c}{MonoFormer} &0.29$\pm$0.002	&0.553$\pm$0.003	&0.755$\pm$0.002	&0.58$\pm$0.003	&2.033$\pm$0.025	&3.18$\pm$0.016	&0.594$\pm$0.003	&0.205$\pm$0.001	&58.692$\pm$0.296

 \\

    \multicolumn{2}{c}{ours} &\textbf{0.903$\pm$0.002}	&\textbf{0.985$\pm$0.002}	&\textbf{0.996$\pm$0.0}	&\uline{0.106$\pm$0.001}	&\textbf{0.098$\pm$0.001}	&\textbf{0.742$\pm$0.01}	&\textbf{0.146$\pm$0.002}	&\uline{0.045$\pm$0.0}	&\uline{13.446$\pm$0.229}

 \\
       \bottomrule
    \end{tabular}}
  \label{tab:SOTA}%
 \vspace{-0.3cm}
\end{table}%
\subsection{Uncertainty Evaluation}
\begin{figure}[htbp]
\setlength{\abovecaptionskip}{0.cm}
\centerline{\includegraphics[width=1\columnwidth]{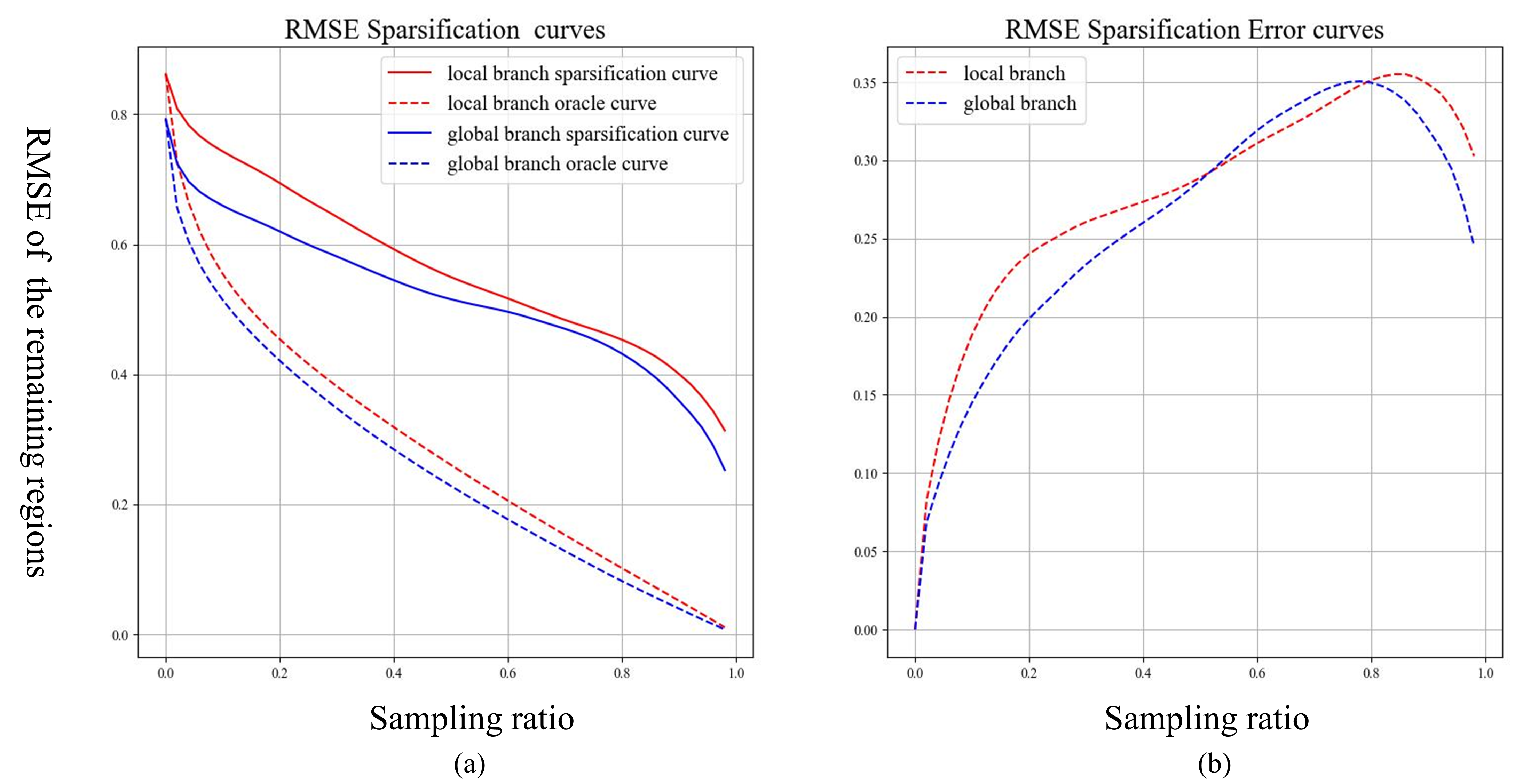}}
\caption{RMSE Sparsification curve of the uncertainty map estimated for the two branches. (a) RMSE Sparsification curve calculated by methods proposed by Matteo et al. The Sparsification curve and Oracle curve are very close in the high uncertainty part (b) RMSE Sparsification error curve is calculated by the subtraction of the Sparsification curve and the Oracle curve. As the error decreases, the gap between the two curves gradually accumulates, but eventually converges to the x-axis.}
\label{fig: uncerteval}
\vspace{-0.7cm}
\end{figure}

Considering that the uncertainty map will directly affect the final results and denote the trustworthy parts of predictions, it is very important to evaluate the quality of the uncertainty map. Here we use the method proposed by Matteo et al.\cite{RN272}. It quantitatively measures the correlation between the uncertainty and the error distribution. In particular, the pixels in the predicted map are sorted in descending order of uncertainty first. Afterwards, the pixels are extracted without replacement, and the RMSE of the remaining region is calculated to plot a Sparsification curve for the RMSE and sampling ratio. The ideal curve named the Oracle curve is obtained by sorting pixels in descending order of RMSE, indicating the error distribution shown by a dotted line in Fig. \ref{fig: uncerteval}.

It can be seen that in the high uncertainty portion of the graph, the depth error and the uncertainty are highly correlated, which means that the local and global branches of our proposed network can both well recognize the high error region of the predicted depth map. As the depth estimation error decreases, the uncertainty estimation error gradually increases but finally decreases and converges to the x-axis.

\subsection{Compared with SOTA Methods}
To further evaluate the performance of the proposed methods, we compare the result of depth estimation with the following state-of-the-art supervised methods. The DenseNet\cite{RN375} and the From-big-to-small methods (BTS)\cite{RN488} are CNN-based methods with specially designed training strategies and prior knowledge. The DepthFormer\cite{RN479} is a Transformer-based model for depth estimation. Furthermore, the TransDepth\cite{RN493} and MonoFormer\cite{bae2023deep} which are CNN-Transformer hybrid models are also included, where MonoFormer is a self-supervised method. In addition, another self-supervised method MonoVit\cite{RN244} is also included in our comparison.
All the methods are trained on synthetic colon data in the EndoSLAM dataset with $16,660$ training images and $1,062$ testing images, which is split according to the experiment design in \cite{RN497}

As shown in Tab.~\ref{tab:SOTA}, our method achieves better results than these SOTA methods on most metrics. In particular, the $\delta_1$, which measures the accuracy of depth estimation, has been improved to \textbf{$0.903$} and the RMSE has been reduced to \textbf{$0.742$}. 

\begin{table*}[htbp]
  \centering
  \caption{Quantitative evaluation results of the model generalization to various out-of-distribution datasets.}
  \resizebox{\linewidth}{!}{
    \begin{tabular}{ccccccccccc|ccccccccc}
    \toprule
     \multicolumn{2}{c}{methods}            & \multicolumn{9}{c}{Small Intestine}                                         & \multicolumn{9}{c}{Stomach} \\
    \midrule
     & & $\delta_1\uparrow$    & $\delta_2\uparrow$    & $\delta_3\uparrow$    & abs\_rel$\downarrow$ & sq\_rel$\downarrow$ & RMSE$\downarrow$  & RMSE\_log$\downarrow$ & log10$\downarrow$ & silog$\downarrow$ & $\delta_1\uparrow$    & $\delta_2\uparrow$    & $\delta_3\uparrow$    & abs\_rel$\downarrow$ & sq\_rel$\downarrow$ & RMSE$\downarrow$  & RMSE\_log$\downarrow$ & log10$\downarrow$ & silog$\downarrow$ \\
    \midrule
    \multicolumn{2}{c}{BTS} &0.968$\pm$0.002	&0.996$\pm$0.002	&\uline{0.998$\pm$0.002}	&0.071$\pm$0.001	&0.292$\pm$0.14	&2.407$\pm$0.049	&0.116$\pm$0.034	&0.033$\pm$0.004	&10.913$\pm$3.273	&0.934$\pm$0.002	&0.981$\pm$0.001	&0.99$\pm$0.001	&0.095$\pm$0.001	&0.218$\pm$0.005	&1.568$\pm$0.018	&0.448$\pm$0.079	&0.058$\pm$0.006	&43.788$\pm$7.804
  \\
    
    \multicolumn{2}{c}{DenseDepth} &0.901$\pm$0.003	&0.985$\pm$0.0	&0.997$\pm$0.0	&0.102$\pm$0.001	&0.369$\pm$0.013	&2.911$\pm$0.072	&0.135$\pm$0.001	&0.042$\pm$0.0	&12.684$\pm$0.145	&0.845$\pm$0.105	&0.94$\pm$0.064	&0.966$\pm$0.037	&0.147$\pm$0.063	&0.503$\pm$0.45	&1.954$\pm$0.668	&0.521$\pm$0.409	&0.119$\pm$0.1	&47.943$\pm$36.38
 \\
    
    \multicolumn{2}{c}{DepthFormer} &\uline{0.969$\pm$0.002}	&0.996$\pm$0.0	&0.999$\pm$0.0	&0.065$\pm$0.003	&\uline{0.209$\pm$0.017}	&\uline{2.325$\pm$0.133}	&0.091$\pm$0.003	&\uline{0.028$\pm$0.001}	&8.502$\pm$0.251	&\textbf{0.964$\pm$0.001}	&\uline{0.992$\pm$0.001}	&\uline{0.996$\pm$0.001}	&0.081$\pm$0.001	&0.193$\pm$0.03	&\uline{1.176$\pm$0.024}	&\uline{0.112$\pm$0.002}	&\uline{0.033$\pm$0.0}	&\uline{10.301$\pm$0.213}
  \\
    
    \multicolumn{2}{c}{TransDepth} &0.969$\pm$0.005	&\uline{0.997$\pm$0.001}	&0.999$\pm$0.0	&\uline{0.065$\pm$0.002}	&0.238$\pm$0.022	&2.629$\pm$0.143	&\uline{0.089$\pm$0.003}	&0.029$\pm$0.001	&\uline{8.292$\pm$0.178}	&0.957$\pm$0.004	&\textbf{0.994$\pm$0.0}	&\textbf{0.998$\pm$0.0}	&\uline{0.08$\pm$0.002}	&\uline{0.168$\pm$0.023}	&1.364$\pm$0.084	&\textbf{0.111$\pm$0.003}	&0.034$\pm$0.001	&\textbf{10.223$\pm$0.223}
  \\
    
    \multicolumn{2}{c}{MonoVit} &0.331$\pm$0.009	&0.604$\pm$0.015	&0.786$\pm$0.012	&0.512$\pm$0.01	&6.162$\pm$0.221	&10.947$\pm$0.25	&0.527$\pm$0.011	&0.184$\pm$0.004	&51.799$\pm$1.107	&0.36$\pm$0.007	&0.641$\pm$0.01	&0.812$\pm$0.008	&0.505$\pm$0.006	&3.068$\pm$0.062	&5.3$\pm$0.12	&0.522$\pm$0.009	&0.18$\pm$0.003	&51.358$\pm$0.898
  \\
    
    \multicolumn{2}{c}{MonoFormer} &0.313$\pm$0.004	&0.573$\pm$0.006	&0.754$\pm$0.007	&0.539$\pm$0.01	&6.903$\pm$0.253	&11.794$\pm$0.178	&0.575$\pm$0.018	&0.197$\pm$0.004	&56.779$\pm$1.83	&0.345$\pm$0.004	&0.612$\pm$0.003	&0.785$\pm$0.002	&0.529$\pm$0.003	&3.474$\pm$0.06	&5.794$\pm$0.07	&0.576$\pm$0.026	&0.192$\pm$0.002	&56.842$\pm$2.606
  \\
    
    \multicolumn{2}{c}{Ours} &\textbf{0.98$\pm$0.001}	&\textbf{0.997$\pm$0.0}	&\textbf{0.999$\pm$0.0}	&\textbf{0.056$\pm$0.001}	&\textbf{0.166$\pm$0.004}	&\textbf{2.141$\pm$0.027}	&\textbf{0.079$\pm$0.002}	&\textbf{0.024$\pm$0.001}	&\textbf{7.379$\pm$0.168}	&\uline{0.961$\pm$0.003}	&0.987$\pm$0.002	&0.993$\pm$0.001	&\textbf{0.075$\pm$0.001}	&\textbf{0.13$\pm$0.006}	&\textbf{1.104$\pm$0.001}	&0.115$\pm$0.003	&\textbf{0.032$\pm$0.001}	&10.645$\pm$0.313
 \\
    \bottomrule
    \toprule
     \multicolumn{2}{c}{methods}            & \multicolumn{9}{c}{Colondepth}                                         & \multicolumn{9}{c}{Scenario} \\
    \midrule
     & & $\delta_1\uparrow$    & $\delta_2\uparrow$    & $\delta_3\uparrow$    & abs\_rel$\downarrow$ & sq\_rel$\downarrow$ & RMSE$\downarrow$  & RMSE\_log$\downarrow$ & log10$\downarrow$ & silog$\downarrow$ & $\delta_1\uparrow$    & $\delta_2\uparrow$    & $\delta_3\uparrow$    & abs\_rel$\downarrow$ & sq\_rel$\downarrow$ & RMSE$\downarrow$  & RMSE\_log$\downarrow$ & log10$\downarrow$ & silog$\downarrow$ \\
    \midrule
       \multicolumn{2}{c}{BTS} &\uline{0.59$\pm$0.066}	&\uline{0.818$\pm$0.075}	&\uline{0.879$\pm$0.071}	&1.269$\pm$0.38	&197.241$\pm$65.187	&37.023$\pm$16.281	&0.662$\pm$0.305	&0.187$\pm$0.091	&62.256$\pm$26.936	&0.653$\pm$0.044	&0.921$\pm$0.022	&\uline{0.98$\pm$0.012}	&0.228$\pm$0.045	&4.946$\pm$3.095	&9.502$\pm$2.391	&0.281$\pm$0.045	&0.089$\pm$0.011	&27.744$\pm$4.524
  \\
    
    \multicolumn{2}{c}{DenseDepth} &0.43$\pm$0.198	&0.613$\pm$0.231	&0.683$\pm$0.223	&1.116$\pm$0.678	&135.531$\pm$107.895	&52.107$\pm$32.26	&1.508$\pm$1.063	&0.458$\pm$0.317	&122.031$\pm$82.229	&0.67$\pm$0.141	&0.82$\pm$0.136	&0.876$\pm$0.107	&0.348$\pm$0.191	&14.4$\pm$11.739	&17.609$\pm$9.662	&0.348$\pm$0.158	&0.114$\pm$0.051	&32.344$\pm$14.006
 \\
    
    \multicolumn{2}{c}{DepthFormer} &0.429$\pm$0.028	&0.695$\pm$0.037	&0.847$\pm$0.031	&\uline{0.398$\pm$0.028}	&\uline{6.104$\pm$0.861}	&\uline{15.182$\pm$1.539}	&\uline{0.458$\pm$0.042}	&\uline{0.157$\pm$0.014}	&\uline{44.727$\pm$4.219}	&\uline{0.708$\pm$0.076}	&\uline{0.925$\pm$0.035}	&0.979$\pm$0.016	&\uline{0.165$\pm$0.028}	&\uline{1.469$\pm$0.508}	&\uline{7.148$\pm$1.269}	&\uline{0.236$\pm$0.039}	&\uline{0.079$\pm$0.014}	&\uline{21.875$\pm$3.803}
  \\
    
    \multicolumn{2}{c}{TransDepth} &0.262$\pm$0.018	&0.531$\pm$0.019	&0.723$\pm$0.01	&0.73$\pm$0.072	&36.992$\pm$16.478	&26.46$\pm$4.013	&0.668$\pm$0.006	&0.234$\pm$0.002	&64.674$\pm$0.241	&0.597$\pm$0.05	&0.842$\pm$0.021	&0.914$\pm$0.006	&0.345$\pm$0.031	&9.048$\pm$1.333	&11.355$\pm$0.621	&0.373$\pm$0.019	&0.117$\pm$0.009	&36.701$\pm$1.893
  \\
    
    \multicolumn{2}{c}{MonoVit} &0.338$\pm$0.005	&0.598$\pm$0.008	&0.771$\pm$0.007	&0.513$\pm$0.008	&9.117$\pm$0.24	&18.01$\pm$0.113	&0.563$\pm$0.006	&0.194$\pm$0.002	&55.596$\pm$0.604	&0.487$\pm$0.022	&0.82$\pm$0.013	&0.938$\pm$0.005	&0.268$\pm$0.011	&2.968$\pm$0.142	&9.744$\pm$0.171	&0.341$\pm$0.009	&0.121$\pm$0.004	&33.704$\pm$0.901
  \\
    
    \multicolumn{2}{c}{MonoFormer} &0.328$\pm$0.0	&0.58$\pm$0.0	&0.749$\pm$0.001	&0.541$\pm$0.009	&11.268$\pm$1.343	&19.002$\pm$0.464	&0.589$\pm$0.006	&0.202$\pm$0.001	&58.213$\pm$0.571	&0.426$\pm$0.007	&0.763$\pm$0.013	&0.915$\pm$0.007	&0.307$\pm$0.008	&3.655$\pm$0.175	&10.475$\pm$0.151	&0.378$\pm$0.007	&0.136$\pm$0.003	&37.494$\pm$0.761
  \\
    
    \multicolumn{2}{c}{Ours} &\textbf{0.71$\pm$0.021}	&\textbf{0.91$\pm$0.007}	&\textbf{0.959$\pm$0.003}	&\textbf{0.282$\pm$0.018}	&\textbf{5.731$\pm$1.554}	&\textbf{8.415$\pm$0.58}	&\textbf{0.273$\pm$0.01}	&\textbf{0.088$\pm$0.003}	&\textbf{26.33$\pm$1.059}	&\textbf{0.836$\pm$0.015}	&\textbf{0.986$\pm$0.002}	&\textbf{0.999$\pm$0.0}	&\textbf{0.124$\pm$0.006}	&\textbf{0.672$\pm$0.043}	&\textbf{4.585$\pm$0.199}	&\textbf{0.161$\pm$0.007}	&\textbf{0.055$\pm$0.003}	&\textbf{15.758$\pm$0.664}
 \\
     \bottomrule
    \end{tabular}}
  \label{tab:stomach_instine}%
\end{table*}

\begin{figure}[htbp]
\setlength{\abovecaptionskip}{0.cm}
\centerline{\includegraphics[width=1\columnwidth]{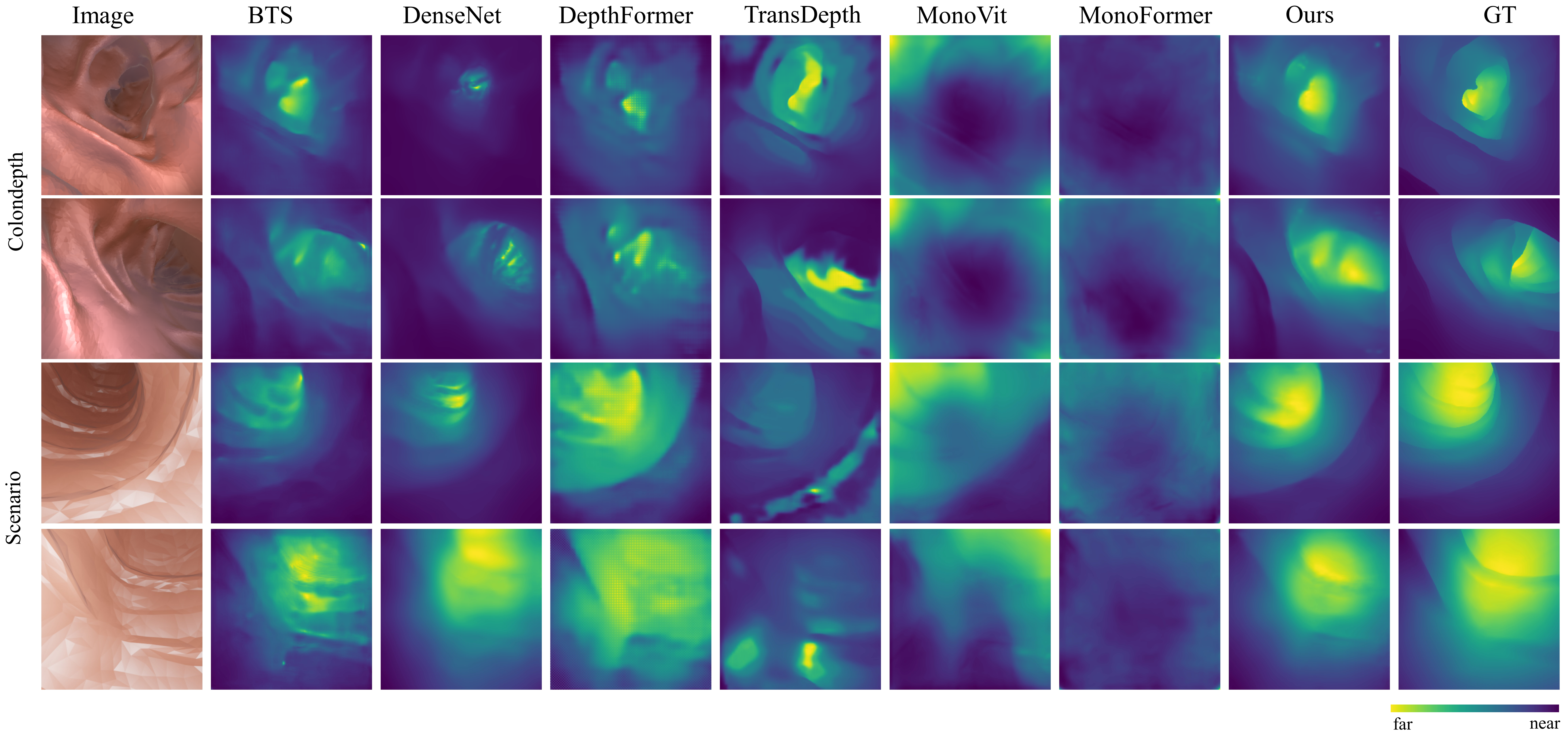}}
\caption{Qualitative comparison of the generalization of models trained on the EndoSLAM dataset to out-of-distribution samples from Colondepth and Scenario datasets. 
Our results strike a balance between obtaining clear boundaries, high structural similarity, and depth uniformity.}
\label{fig:generalization}
\vspace{-0.8cm}
\end{figure}

\subsection{Generalizing to Out-of-distribution Samples}
Good performance on a single simulated dataset is inadequate due to the distribution gap between simulated and real data. It is therefore crucial to assess the model's ability to generalize to out-of-distribution data.
    However, depth ground-truth for real data is currently unavailable due to the difficulty of data acquisition, making quantitative depth estimation evaluation on real data is currently not feasible.
    Considering that conducting quantitative evaluations on a large number of datasets from different distributions can also demonstrate our model's generalization ability to out-of-distribution data, serving as a reference for our model's generalization ability to real data, we first conducted two types of quantitative evaluations: generalization to different anatomical structures and generalization to different datasets.
    After that, we qualitatively test their performance on the clinical dataset and conduct 3D reconstruction based on the predicted depth maps to illustrate the superiority  of our results.

In Tab.~\ref{tab:stomach_instine}, we trained the model with the simulated colon data in the EndoSLAM dataset and tested the model on the simulated stomach images and small intestine images. The test set of the stomach and small intestine contains $1,548$ and $1,257$ images, respectively. As shown in Tab.~\ref{tab:stomach_instine}, our method achieved the best results for most of the evaluation metrics, while a few indicators reached suboptimal results close to the optimal ones. We also perform a qualitative comparison of the depth estimation results of different anatomical structure from the EndoSLAM dataset in Fig. \ref{fig:endoslam}. Compared with other methods, our method can reconstruct more detailed depth maps without being affected by noise such as reflective regions.

Our method's superiority is further validated on the test sets of both the Colondepth dataset (consisting of $1,092$ images) and the Scenario dataset (comprising $100$ images). Our method achieves SOTA results on all the indicators on both of the datasets. 
The visual comparison in Fig. \ref{fig:generalization} shows that our method can achieve clear boundaries in the depth maps while ensuring the structure similarity and depth uniformity of the depth maps.

\begin{figure*}[htbp]
\setlength{\abovecaptionskip}{0.cm}
\centerline{\includegraphics[width=1.7\columnwidth]{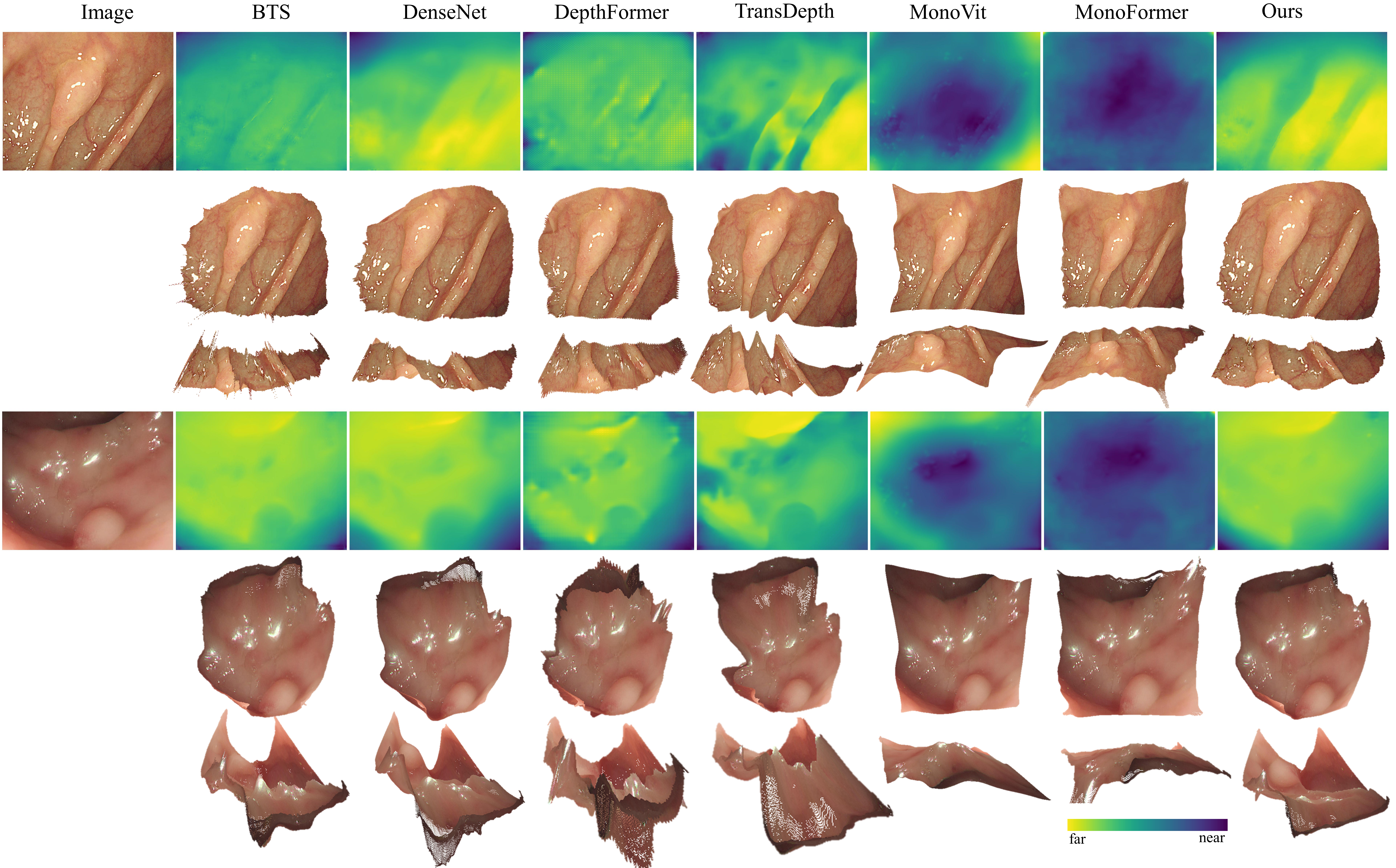}}
\caption{ 
The predicted depth and the 3D reconstruction results of different methods on the real images with polyps.
Our method can clearly reflect the boundaries of polyps while ensuring uniform depth estimation in other regions. The depth estimation results of polyps will not be affected by reflections, air bubbles and different camera views, which are the various challenges affecting colon depth estimation. }
\label{fig:real}
\vspace{-0.7cm}
\end{figure*}

For clinical applications, we employed the RenjiVideoDB dataset\cite{RN491} which was collected during actual examinations for polyp detection and classification study. 
The depth estimation results are shown in Fig.~\ref{fig:real}. For clinical images that contain polyps, reflective regions, and air bubbles, our network can predict a clear edge of the polyp while predicting a smooth depth map for other regions in the image despite various camera views.
Additionally, we have conducted 3D reconstruction from the images and the predicted depth maps. The reconstruction results also demonstrate that our 3D images can reflect more reasonable colon structures.
The depth map obtained by our method can reflect the presence of polyps in clinical colonoscopy images, which implies our method can be applied to a real polyp detection and classification task.

\section{Discussion}
Our study proposed a novel network architecture containing a CNN-based local branch and a Transformer-based global branch to deal with the errors caused by various noises and complex lighting conditions during colonoscopy depth estimation. To make full use of the complementary information, we propose a novel fusion module based on uncertainty estimation. The fusion module can selectively combine the trustworthy prediction parts of the two branches based on the uncertainty map to construct a more accurate depth map.

To the best of our knowledge, it is the first study to adapt the CNN-Transformer hybrid model with uncertainty estimation to colonoscopy depth estimation. Our model achieves SOTA results on the EndoSLAM dataset and has an excellent generalization ability for a variety of out-of-distribution images. 
It is worth mentioning that our network can obtain reasonable depth estimation maps on raw clinical colonoscopy images with polyps. The edges of the polyps are clearer in the depth map and the depth estimation is not affected by reflective regions that are not eliminated by preprocessing. This is strong evidence that our depth estimation method can well serve as a basis for various clinical tasks, such as polyp detection and classification.

We believe that the generalization improvement may be associated with the uncertainty-based fusion module we proposed. The uncertainty map generated from the local branch gives higher weights to the regions of reflection, while the uncertainty map derived from the global branch pays more attention to the edges.
These regions are the main and common challenge for depth estimation for various colonoscopy images. Our uncertainty-based fusion module allows the network to leverage the complementary advantages of CNN and Transformer and obtain a reasonable depth map with information about the probability distribution of errors when faced with out-of-distribution samples and therefore has an excellent generalization ability.

Although our study has significantly improved model generalization, it is important to acknowledge several limitations. 
Firstly, this study focuses solely on estimating depth maps from individual monocular colonoscopy images, overlooking the valuable spatio-temporal information in the entire video sequence. Incorporating this information in future work could further optimize our approach.
Secondly, quantitative evaluation of generalization on real-world data should be performed on clinical datasets with available ground-truth depth data.
In addition, the hybrid network is inefficient for some applications, so the lightweight network design for two branches is needed to be explored continuing the ideas of Yang et al.\cite{RN492}.
We hope that addressing these limitations in future work will further enhance the performance of the model.

\section{Conclusion}

In this paper, we proposed a novel CNN-Transformer hybrid network to handle the noise of reflective regions, unclear boundaries, and regions with illumination variation in the monocular colonoscopy depth estimation task. An uncertainty-based fusion model is used to merge the complementary global-local information which is captured by the two branches. Our network achieves SOTA results on the EndoSLAM dataset and have excellent generalization performance to out-of-distribution data, including different anatomical structures, simulated datasets, and real clinical datasets. This method holds promise in providing support for various real clinical tasks, encompassing polyp detection, polyp size classification, and endoscopic navigation.

\section{Acknowledgment}
The authors would like to thank Yongming Yang from the Shenyang Institute of Automation, Chinese Academy of Sciences, for his valuable suggestions for results evaluation and visualization.

\end{document}